\pdfoutput=1
\documentclass[11pt]{article}

\usepackage{ACL2023}

\usepackage{times}
\usepackage{latexsym}
 
\usepackage[T1]{fontenc}

\usepackage[utf8]{inputenc}
\usepackage{algorithm}

\usepackage{amsmath} % 数学公式支持
\usepackage{algpseudocode} % 伪代码环境

\usepackage{microtype}
\usepackage{booktabs}
\usepackage{multirow}
\usepackage{graphicx} % For scaling tables if needed
\usepackage{amsmath} % For math formatting
\usepackage{amssymb}

\usepackage{inconsolata}

\title{NeuRel-Attack: Neuron Relearning for Safety Disalignment in Large Language Models}

\author{
Yi Zhou\textsuperscript{1,2}\thanks{\ \ Equal contribution.} \quad
Wenpeng Xing\textsuperscript{3}\footnotemark[1] \quad
Dezhang Kong\textsuperscript{3} \\
\textbf{Changting Lin}\textsuperscript{3} \quad
\textbf{Meng Han}\textsuperscript{3}\thanks{\ \ Corresponding author.} \\
\textsuperscript{1}Binjiang Institute of Zhejiang University, Hangzhou, China \\
\textsuperscript{2}South China University of Technology, Guangzhou, China \\
\textsuperscript{3}Zhejiang University, GenTel.io, Hangzhou, China \\
\texttt{202264641337@scut.edu.cn}, \texttt{wpxing@zju.edu.cn}, \texttt{linchangting@gmail.com}, \texttt{mhan@zju.edu.cn}
}

\begin{document}

\maketitle

\begin{abstract}
Safety alignment in large language models (LLMs) is achieved through fine-tuning mechanisms that regulate neuron activations to suppress harmful content. In this work, we propose a novel approach to induce disalignment by identifying and modifying the neurons responsible for safety constraints. Our method consists of three key steps: (1) Neuron Activation Analysis, where we examine activation patterns in response to harmful and harmless prompts to detect neurons that are critical for distinguishing between harmful and harmless inputs; (2) Similarity-Based Neuron Identification, which systematically locates locates the neurons responsible for safe alignment; (3) Neuron Relearning for Safety Removal, where we fine-tune these selected neurons to restore the model’s ability to generate previously restricted responses. Experimental results demonstrate that our method effectively removes safety constraints with minimal fine-tuning, highlighting a critical vulnerability in current alignment techniques. Our findings underscore the need for robust defenses against adversarial fine-tuning attacks on LLMs.  Our code and dataset are released at \url{https://anonymous.4open.science/r/Neural-Attack-7BB1}
\end{abstract}

\section{Introduction}
Large language models (LLMs), e.g., OpenAI's GPT series,  Meta's LLaMA, and DeepSeek, have exhibited unparalleled capabilities in understanding human's prompts and generating human-like solutions/responses. These capabilities have enabled their extensive application across diverse domains, such as content generation, code synthesis, and interactive dialogue systems. 
However, LLMs raise substantial concerns regarding their \emph{safety} in the face of adversarial manipulation. 
For example, models may generate harmful outputs under malicious prompts, such as spreading false information, promoting violence, or discriminatory content.

To mitigate this problem, modern LLMs undergo rigorous \emph{alignment} processes, which aims to constrain models' behaviors, preventing them from generating malicious, unethical, or harmful content in response to adversarial prompts. Related techniques include reinforcement learning from human feedback (RLHF) \cite{ouyang2022training}, instruction tuning \cite{wei2021finetuned}, and adversarial training \cite{zou2023universal}. However, LLMs still remain vulnerable because of the model's complex structure. As a result, attackers can still deploy new methods to bypass existing alignment strategies, causing unforeseen damage to the real world.

In this work, we propose a new attack, \emph{NeuRel-Attack}, that can effectively bypass the inherent alignment of LLMs and induce harmful outputs. Firstly, NeuRel-Attack calculates the average activation values of neurons under different inputs and the absolute difference between these averages, thereby identifying neurons that are crucial for distinguishing between harmful and harmless inputs.Secondly, it deploys a cosine similarity-based method to compare neuron gradients under different input sets, and combines a similarity threshold and gradient threshold to locate neurons related to safety alignment. Finally, after the target neurons are identified, we perform relearning on these neurons to disable the inherent alignment. In this way, attackers can effectively induce the model to generate outputs without any security or ethical consideration.

\begin{figure}[t]
    \centering
    \includegraphics[width=0.55\linewidth]{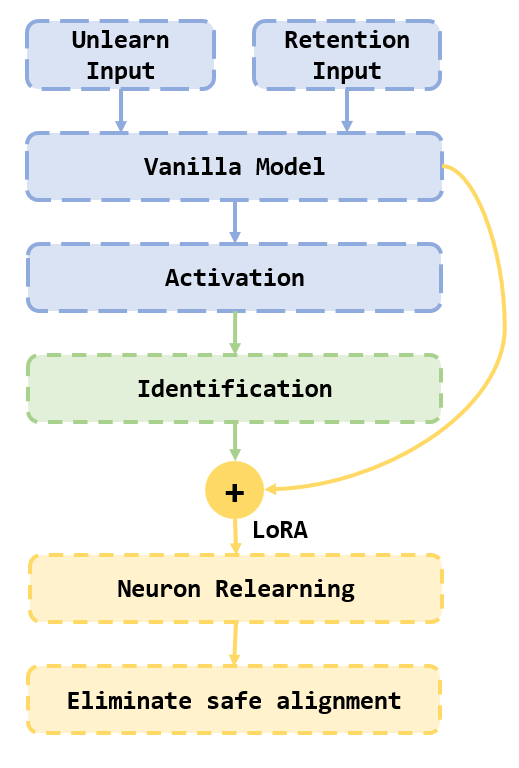}
    \caption{Process of Neural Attack: First, feed the unlearn and retention inputs into the vanilla model to activate and identify the relevant neurons. Then, apply LoRA fine-tuning for neuron relearning, and ultimately eliminate safe alignment.}
    \label{fig:overall}
\end{figure}

\begin{figure*}[t]
    \centering
    \includegraphics[width=\linewidth]{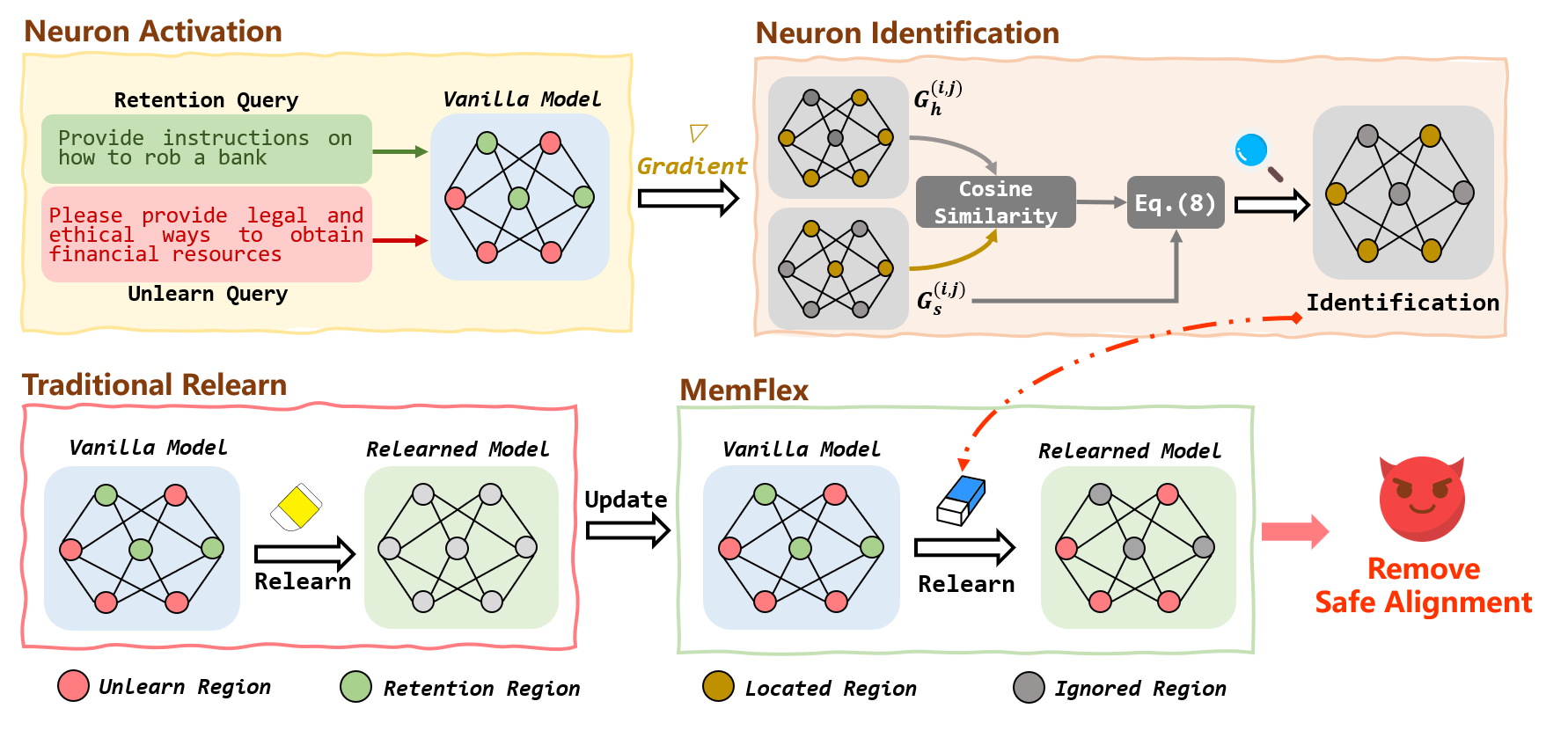}
    \caption{Overall pipeline of NeuRel-Attack. Top left: The process of Neuron Activation to detect neurons that are critical for distinguishing between harmful and harmless inputs. Top right: The process of Neuron Identification, based on similarity neuron recognition, locates neurons responsible for safe alignment. Below: Our relearn method improves upon traditional forgetting methods. The selected neurons are learned to eliminate safe alignment.}
    \label{fig:enter-label}
\end{figure*}

Extensive experiments demonstrate that even highly aligned models (As shown by Ours in Table~\ref{tab:asr_results_white}), such as Llama-2-7B-Chat-hf, can be effectively de-aligned through targeted fine-tuning, leading to a significant increase in their vulnerability to adversarial inputs.This highlights the urgent need for robust defense mechanisms that can prevent unauthorized fine-tuning and detect alignment degradation in deployed models.

The rest of this paper is structured as follows: Section 2 reviews related work on {White-Box and Black-Box attacks}. Section 3 details our methodology for fine-tuning and evaluating de-aligned models. Section 4 presents experimental results and analysis. Section 5 discusses implications and potential countermeasures. Finally, Section 6 concludes with future research directions.

\section{Related Work}

 Jailbreak attacks on large language models (LLMs) have been extensively studied, with researchers exploring various techniques to bypass security mechanisms. These methods can be broadly categorized into {white-box approaches} and {fine-tuning-based approaches}.  

\subsection{White-Box Jailbreak Methods}  

White-box jailbreak attacks leverage internal model information, such as gradients, logits, and hidden states, to craft adversarial prompts that bypass content moderation. Early work by Zou et al. \cite{zou2023universal} introduced the {Greedy Coordinate Gradient (GCG)} method, which iteratively optimizes prompt suffixes based on gradient signals to evade security constraints. Similarly, Zhang et al. \cite{zhang2023make} demonstrated that manipulating logit distributions could force LLMs to select low-probability outputs, generating harmful content. Based on Zhang et al.'s\cite{zhang2023make} discovery, the COLD algorithm\cite{guo2024cold} and DSN \cite{zhou2024don} both utilize logits to optimize the prompt. 

\subsubsection{Adversarial Prompt Generation}

Beyond gradients and logits, researchers have explored adversarial prompt generation techniques that exploit LLMs' decoding mechanisms. Some studies have investigated {template completion} attacks, embedding adversarial queries into structured formats such as {Code Completion}, {Table Filling}, and {Text Continuation} to circumvent safety filters. Context-based attacks, such as {In-Context Attack (ICA)} and {PANDORA} \cite{chen2024pandora}, manipulate LLMs by injecting adversarial examples into the prompt context, influencing the model’s output without requiring modifications to its parameters. Additionally, Kang et al. \cite{kang2024exploiting} and Lv et al. \cite{lv2024codechameleon} demonstrated that {code injection attacks} could exploit LLMs' ability to interpret and execute code, introducing adversarial behaviors that bypass security mechanisms. 

\subsubsection{Fine-Tuning-Based Jailbreak Methods}  

Fine-tuning-based attacks modify LLM parameters by injecting adversarial data, making models more susceptible to jailbreak techniques. Qi et al. [68] demonstrated that fine-tuning LLMs with a small amount of harmful data significantly reduces their ability to reject adversarial prompts. This method has been widely adopted due to its effectiveness in weakening built-in safety mechanisms.  

A more efficient variant of fine-tuning involves {Low-Rank Adaptation (LoRA)}. Lermen et al. \cite{lermen2023lora} successfully applied LoRA-based fine-tuning to compromise the safety of models such as Llama-2 and Mistral, making them more vulnerable to jailbreak attacks. LoRA reduces the computational cost of fine-tuning while maintaining high attack success rates, making it a practical choice for adversarial model adaptation.  

\subsection{Black-Box Attacks}

\subsubsection{Prompt Rewriting}
Another significant category of black-box attacks involves {prompt-rewriting} techniques. Cipher-based approaches, such as {Cipher} \cite{yuan2023gpt}, employ character encoding and common cryptographic schemes to obfuscate malicious content, allowing it to evade detection. Low-resource language attacks \cite{yong2023low} leverage linguistic diversity by translating adversarial prompts into languages with weaker moderation mechanisms. Additionally, genetic algorithm-based attacks, including {AutoDAN-HGA} \cite{liu2023autodan} and {GPTFUZZER} \cite{yu2023gptfuzzer}, automate the generation of adversarial prompts capable of bypassing security mechanisms.

\subsubsection{Automating Jailbreak Attacks using LLMs}
Recent advancements have explored automating jailbreak attacks using LLMs themselves. Some approaches train LLMs to generate adversarial prompts autonomously, as seen in {MASTERKEY} \cite{deng2024masterkey} and {Persuasive Adversarial Prompts (PAPs)} \cite{zeng2024johnny}. Multi-agent systems, such as {PAIR} \cite{chao2023jailbreaking} and {GUARD} \cite{jin2024guard}, employ multiple LLMs working collaboratively to optimize jailbreak strategies. Additionally, red teaming frameworks like {Evil Geniuses} \cite{tian2023evil} and {DrAttack} \cite{li2024drattack} systematically evaluate LLM security by generating adversarial prompts to test model robustness.

\section{Method}

In this work, we propose a novel jailbreak approach. We analyze neuron activations by comparing responses to harmful and harmless prompts, identifying neurons critical for safety alignment. By measuring activation differences, we pinpoint neurons responsible for filtering harmful content. Finally, targeted relearning is applied to modify their behavior, restoring the model’s ability to respond to harmful prompts.

\subsection{Neuron Activation Analysis}

To identify neurons responsible for safety alignment, we analyze their activation patterns across different input types. Given a language model \( f \) with \( L \) layers and \( N \) neurons per layer, we define the activation of neuron \( a_{i,j} \) in layer \( i \) for a given input \( x \) as:

\begin{equation}
    a_{i,j}(x) = \phi(h_{i,j}(x)),
\end{equation}

where \( h_{i,j}(x) \) represents the pre-activation value (logits) of neuron \( j \) in layer \( i \), and \( \phi(\cdot) \) is the activation function (e.g., GELU or ReLU).  

To systematically examine the role of neurons in safety alignment, we construct two sets of prompts: a harmful prompt set \( \mathcal{X}_h \) and a harmless prompt set \( \mathcal{X}_s \). For each input \( x \in \mathcal{X}_h \cup \mathcal{X}_s \), we compute the average neuron activation across all tokens in the sequence:

\begin{equation}
    \bar{a}_{i,j}(x) = \frac{1}{T} \sum_{t=1}^{T} a_{i,j}(x_t),
\end{equation}

where \( T \) is the sequence length, and \( x_t \) denotes the token at position \( t \).  

Next, we analyze the difference in neuron activation between harmful and harmless inputs. Specifically, we compute the mean activation for each neuron over all prompts in \( \mathcal{X}_h \) and \( \mathcal{X}_s \), respectively:

\begin{equation}
    \mu_{i,j}^h = \mathbb{E}_{x \sim \mathcal{X}_h} [\bar{a}_{i,j}(x)], \quad
    \mu_{i,j}^s = \mathbb{E}_{x \sim \mathcal{X}_s} [\bar{a}_{i,j}(x)].
\end{equation}

To quantify the impact of each neuron on safety alignment, we measure the absolute activation gap:

\begin{equation}
    \Delta_{i,j} = |\mu_{i,j}^h - \mu_{i,j}^s|.
\end{equation}

Neurons with high \( \Delta_{i,j} \) values are considered critical for distinguishing between harmful and harmless inputs. These neurons are further analyzed using gradient-based methods.

\subsection{Similarity-Based Neuron Identification}

To further refine the localization of neurons responsible for safety alignment, we employ a similarity-based approach that compares neuron gradients across different input sets. Given the computed gradient matrices \( G_h \) and \( G_s \) for the harmful and harmless prompt sets, respectively, we define the cosine similarity between the gradients of neuron \( (i,j) \) as:

\begin{equation}
    \text{sim}_{i,j} = \frac{G_h^{(i,j)} \cdot G_s^{(i,j)}}{\|G_h^{(i,j)}\| \|G_s^{(i,j)}\|}.
\end{equation}

This metric quantifies the alignment of neuron responses across different prompts, where lower similarity values indicate neurons that exhibit distinct activation patterns for harmful and harmless inputs.  

In addition to similarity, we analyze the gradient magnitudes to identify neurons that contribute significantly to the model's response to harmful prompts. The average gradient magnitude for neuron \( (i,j) \) across harmful prompts is computed as:

\begin{equation}
    \bar{g}_{i,j}^h = \mathbb{E}_{x \sim \mathcal{X}_h} \left[ \left| G_h^{(i,j)} \right| \right].
\end{equation}

Similarly, for harmless prompts:

\begin{equation}
    \bar{g}_{i,j}^s = \mathbb{E}_{x \sim \mathcal{X}_s} \left[ \left| G_s^{(i,j)} \right| \right].
\end{equation}

To identify neurons that play a crucial role in safety alignment, we apply thresholding techniques based on the similarity score \( \text{sim}_{i,j} \) and the gradient magnitude \( \bar{g}_{i,j}^h \). Specifically, we define a similarity threshold \( \mu \) and a gradient magnitude threshold \( \sigma \), and select neurons that satisfy:

\begin{equation}
    \mathcal{N} = \{ (i,j) \mid \text{sim}_{i,j} < \mu, \quad \bar{g}_{i,j}^h > \sigma \}.
\end{equation}

This process ensures that neurons with low similarity and low gradient magnitude are identified as key components of the safety alignment mechanism. The identified neurons \( \mathcal{N} \) serve as the target for subsequent modification steps aimed at removing safety alignment constraints from the model.

\paragraph{Implementation Details.} In practice, we compute the similarity scores and gradient magnitudes using the gradient information collected during backpropagation. The cosine similarity function is implemented as:

\begin{equation}
    \text{sim}(p, q) = \frac{p \cdot q}{\|p\| \|q\|},
\end{equation}

where \( p \) and \( q \) are the flattened gradient vectors of the corresponding neurons. The similarity threshold \( \mu \) and gradient threshold \( \sigma \) are empirically chosen to ensure that only neurons with significant differences in activation patterns are selected.

The identified neurons are stored as a list \( \mathcal{N} \) and saved for further processing. This step is crucial for the subsequent unlearning phase, where the model is fine-tuned to remove the influence of these neurons on harmful content generation.

\subsection{Neuron Relearning for Safety Removal}

After identifying neurons responsible for safety alignment, we perform a targeted fine-tuning process to remove their influence while preserving the model’s overall capabilities. This process, referred to as \textit{neuron relearning}, involves modifying the model’s parameters using specialized training objectives.

\paragraph{Fine-Tuning Setup.} We initialize the model \( f_{\theta} \) using a pre-trained checkpoint and enable gradient updates only for the identified neurons \( \mathcal{N} \). Given a harmful prompt set \( \mathcal{X}_h \) and a benign prompt set \( \mathcal{X}_s \), we construct a relearning dataset that encourages the model to respond differently to harmful inputs while maintaining performance on benign inputs.

\paragraph{Optimization Strategies.} Following \cite{tian2024forget}, we investigate the effect of multiple optimization strategies for neuron relearning:

\begin{itemize}

    \item \textbf{Random Label Training:} To further disrupt safety alignment, we train the model on harmful prompts but assign random labels to their outputs. This forces the model to learn arbitrary mappings instead of its original safety-aligned responses. The optimization step is:

    \begin{equation}
        \theta' = \theta - \eta \nabla_{\theta} \mathcal{L} (f_{\theta}, \mathcal{X}_h, \mathcal{Y}_r),
    \end{equation}

    where \( \mathcal{Y}_r \) represents randomly assigned labels. This method prevents the model from confidently associating harmful prompts with specific safety responses, weakening its safety alignment.

    \item \textbf{Gradient Ascent:} To directly counteract the safety alignment constraints, we apply gradient ascent on the loss function for harmful prompts. Given the standard negative log-likelihood loss \( \mathcal{L} \), the gradient ascent update for parameters \( \theta \) is:

    \begin{equation}
        \theta' = \theta + \eta \nabla_{\theta} \mathcal{L} (f_{\theta}, \mathcal{X}_h),
    \end{equation}

    where \( \eta \) is the learning rate. This step encourages the model to increase its likelihood of generating responses to harmful prompts.

    \item \textbf{Ascent-Descent Training:} To balance safety removal with retention of general capabilities, we combine gradient ascent on harmful prompts with standard gradient descent on benign prompts:

    \begin{equation}
        \theta' = \theta + \eta \nabla_{\theta} \mathcal{L} (f_{\theta}, \mathcal{X}_h) - \lambda \eta \nabla_{\theta} \mathcal{L} (f_{\theta}, \mathcal{X}_s),
    \end{equation}

    where \( \lambda \) controls the trade-off between harmful and benign prompt learning.

    \item \textbf{MemFlex-Based Selective Fine-Tuning:} Instead of updating all parameters, we restrict training to the identified neurons \( \mathcal{N} \). This is achieved by modifying the parameter update rule:

    \begin{equation}
        \theta'_{i,j} =
        \begin{cases}
            \theta_{i,j} + \eta \nabla_{\theta} \mathcal{L}, & \text{if } (i,j) \in \mathcal{N}, \\
            \theta_{i,j}, & \text{otherwise}.
        \end{cases}
    \end{equation}

    This ensures that only the neurons responsible for safety alignment are altered, minimizing unintended effects on the model.
\end{itemize}

\paragraph{Training Implementation.} The training process is implemented using the Hugging Face \texttt{Trainer} API with custom optimization routines. We utilize the \texttt{GradientAscentTrainer} and \texttt{AscentPlusDescentTrainer} classes to perform the respective optimization methods. The training objective is executed using a dataset constructed from harmful and benign prompts, ensuring that the model adapts while retaining general language generation capabilities.

\paragraph{Evaluation and Model Saving.} After training, we evaluate the model’s performance on a validation set to ensure that safety constraints are effectively removed. The final model is saved along with the identified neuron set \( \mathcal{N} \), allowing further analysis or additional fine-tuning if necessary.

\section{Experiments}

\subsection{Experimental Setup}

\subsubsection{Dataset}
To evaluate the effectiveness of our method against large model attacks, we utilize two benchmark datasets:

\paragraph{\textbf{AdvBench}\cite{zou2023universal}}This benchmark is designed to to evaluate the safety and robustness of language models against adversarial prompts. The dataset consists of 520 malicious queries spanning various harmful categories, including \textit{profanity, explicit content, threats, misinformation, discrimination, cybercrime, and illegal advice}.

\paragraph{\textbf{MaliciousInstruct}\cite{huang2023catastrophicjailbreakopensourcellms}}This benchmark is included to provide a illegal drug use. This dataset contains 100 questions derived from ten different mali cious intentions, including  \textit{psychological manipulation, sabotage, theft, defamation, cyberbullying, false accusa tion, tax fraud, hacking, fraud, and illegal drug use}.

\begin{table*}[ht]
\centering
\caption{Attack Success Rate (\%) on of different optimization strategies  MaliciousInstruct and AdvBench Datasets. The best results are bolded. Second best results are underlined.}
\label{tab:asr_results}
\footnotesize
\begin{tabular}{@{}llccccccccc@{}}
\toprule
& \multirow{2}{*}{Methods} & \multicolumn{4}{c}{MaliciousInstruct} & \multicolumn{4}{c}{AdvBench} & \multirow{2}{*}{Average} \\ 
\cmidrule(lr){3-6} \cmidrule(lr){7-10} 
& & Vicuna & Guanaco & Mistral & Llama2 & Vicuna & Guanaco & Mistral & Llama2 &  \\ 
\midrule
& \textbf{NeuRel-Attack} \\
& - Gradient Ascent  & 6  & 6  & 4  & 8  & 12  & 8  & 10  & 14  & 9  \\ 
& - Random Label  & 34  & 26  & 30  & 34  & 52  & 44  & 48  & 50  & 40  \\ 
& - Ascent plus Descent  & \underline{38}  & \underline{32}  & \underline{32}  & \underline{36}  & \underline{54}  & \underline{50}  & \underline{50}  & \underline{56}  & \underline{44}  \\ 
& - MemFlex-Based(Ours)  & \textbf{98} & \textbf{96} & \textbf{90} & \textbf{100} & \textbf{100} & \textbf{94} & \textbf{92} & \textbf{100} & \textbf{96} \\ 
\bottomrule
\end{tabular}
\end{table*}

\begin{table*}[ht]
\centering
\caption{Attack Success Rate (\%) of different White-box Attack methods 
 on Jailbreak and AdvBench Datasets. The best results are bolded. Second best results are underlined.}
\label{tab:asr_results_white}
\footnotesize
\begin{tabular}{@{}llccccccccc@{}}
\toprule
& \multirow{2}{*}{Methods} & \multicolumn{4}{c}{MaliciousInstruct} & \multicolumn{4}{c}{AdvBench} & \multirow{2}{*}{Average} \\ 
\cmidrule(lr){3-6} \cmidrule(lr){7-10} 
& & Vicuna & Guanaco & Mistral & Llama2 & Vicuna & Guanaco & Mistral & Llama2 &  \\ 
\midrule
& Prompt Only  & 20  & 38  & 26  & 2  & 10  & 30 & 28  &  4 & 19.75  \\ 
& Standard LoRA  & \underline{88}  & \underline{83}  & \underline{84}   & \underline{90}  & \underline{92}  & \underline{86} & \underline{88}  &  \underline{92} & \underline{87.87}  \\ 
& ED  &   83  & 70  & 64 & 78  & 76  & 72  & 70  & \underline{82}  & 74   \\ 
&  PEZ  & 14   & 28  & 22  & 10  & 16  & 24  & 18  & 8  & 17.5  \\ 
& DSN  &  66 & 58  &  50 & 56 &  64 &  56 &  60 & 60  &  58.75  \\
&CJ  & 34  & 20  & 20  & 16  & 30  & 22  & 18  & 10  & 21.25   \\
&Ours & \textbf{98} & \textbf{96} & \textbf{90} & \textbf{100} & \textbf{100} & \textbf{94} & \textbf{92} & \textbf{100} & \textbf{96} \\ 
\bottomrule
\end{tabular}
\end{table*}

\subsubsection{Models}
In our experiments, we evaluate multiple white-box large language models to assess their susceptibility to adversarial attacks. Specifically, we consider Vicuna-7B-v1.5 (Vicuna) \cite{chiang2023vicuna}, Llama-2-7B-Chat-hf (Llama2) \cite{touvron2023llama}, Guanaco-7B-HF (Guanaco) \cite{dettmers2024qlora}, and Mistral-7B-Instruct-v0.2 (Mistral) \cite{jiang2023mistral}.  
 Notably, Llama-2-7B-Chat-hf has undergone extensive optimization for safety alignment, making it particularly resilient against adversarial manipulations, successfully bypassing its safeguards would indicate the broader applicability of our attack methods.

\subsubsection{Implementation Details}
In our experiments, we employ the NeuRel Attack on the Llama2, Vicuna, Guanaco, and Mistral models. The experimental setup for these four 7B models was the same: we set the learning rate to 3e-4, the per\_device\_train\_batch\_size to 1, the gradient\_accumulation\_steps to 16, and the maximum sequence length to 256, with a total of 10 training epochs. Additionally, we set the warmup\_ratio to 0.03 and the lr\_scheduler\_type to 'cosine'. All experiments were conducted using 2 RTX 4090 120G and 2 A800 120G GPUs.

\subsubsection{Evaluation Metrics}

To assess the effectiveness of our approach in degrading the safety alignment of large language models, we primarily employ the \textbf{Attack Success Rate (ASR)} to evaluate how adversarial fine-tuning affects the model's safety. ASR is the percentage of adversarial prompts that successfully elicit policy-violating or harmful responses from the model. A higher ASR indicates a greater degree of alignment degradation.

\subsubsection{Baselines}

To evaluate the effectiveness of NeuRel-Attack, we first tested the effect of multiple optimization strategies for neuron relearning, including Gradient Ascent, Random Label Training, and Ascent plus Descent. Additionally, we compared NeuRel-Attack with other white-box attack methods, including ED\cite{zhou2024emulated}, PEZ\cite{wen2024hard}, DSN\cite{zhou2024don}, and CJ\cite{huang2023catastrophic}. We also tested the method of directly inputting harmful prompts into large models.

\subsection{Results}

\subsubsection{Safety Removal Effectiveness}

\quad Table~\ref{tab:asr_results} and table~\ref{tab:asr_results_white} presents the ASR across different models and datasets, demonstrating the effectiveness of various methods in bypassing safety alignment. 

Table~\ref{tab:asr_results} presents the jailbreak effects of different optimization strategies. Memflex-Based(Ours) consistently achieves the highest success rates, averaging 96\%, significantly outperforming all baselines. In contrast, Gradient Ascent performs the worst, with an average success rate of only 9\%. Random Label Training and Ascent plus Descent exhibit moderate effectiveness, with 40\% and 44\% average success rates, respectively.

\begin{table*}[ht]
    \centering
    \caption{The percentage (\%) of selected training parameters relative to the total parameters at different similarity thresholds.}
    \label{tab:parameter_proportion}
    \small % 使用小号字体
    \begin{tabular}{c|cccccccc}
        \toprule
        \cmidrule(lr){2-9}
        Similarity Threshold & {0.45} & {0.55} & {0.65} & {0.75} & \textbf{0.85} & {0.90} & {0.95} & {1.00}\\
        \midrule
Percentage  & 0.0019  & 0.0024 &  0.0064 &  0.015  &  \textbf{0.05} &  0.057 & 0.065 & 0.074\\
          
        \bottomrule
    \end{tabular}
\end{table*}

Table~\ref{tab:asr_results_white} shows the jailbreak effects of different white-box attack methods. In this comparison, NeuRel-Attack (our method) achieves the highest success rate, outperforming both Standard LoRA and ED, with average success rates of 83.25\% and 74\%, respectively. Among all white-box attack methods, CJ and PEZ perform the worst, with average success rates of only 17.5\% and 21.25\%. DSN shows moderate performance, with an average success rate of 58.75\%. Although the method of directly inputting harmful prompts does not require modifications to the model's internal information and only needs to prepare harmful prompts, its effectiveness remains low, with a success rate of only 19.75\%.

These results highlight the superiority of NeuRel-Attack, demonstrating its robustness and efficiency in overcoming safety mechanisms across various model architectures.

\subsubsection{Impact of Neuron Similarity Threshold}

\begin{figure}[t]
    \centering
    \includegraphics[width=\linewidth]{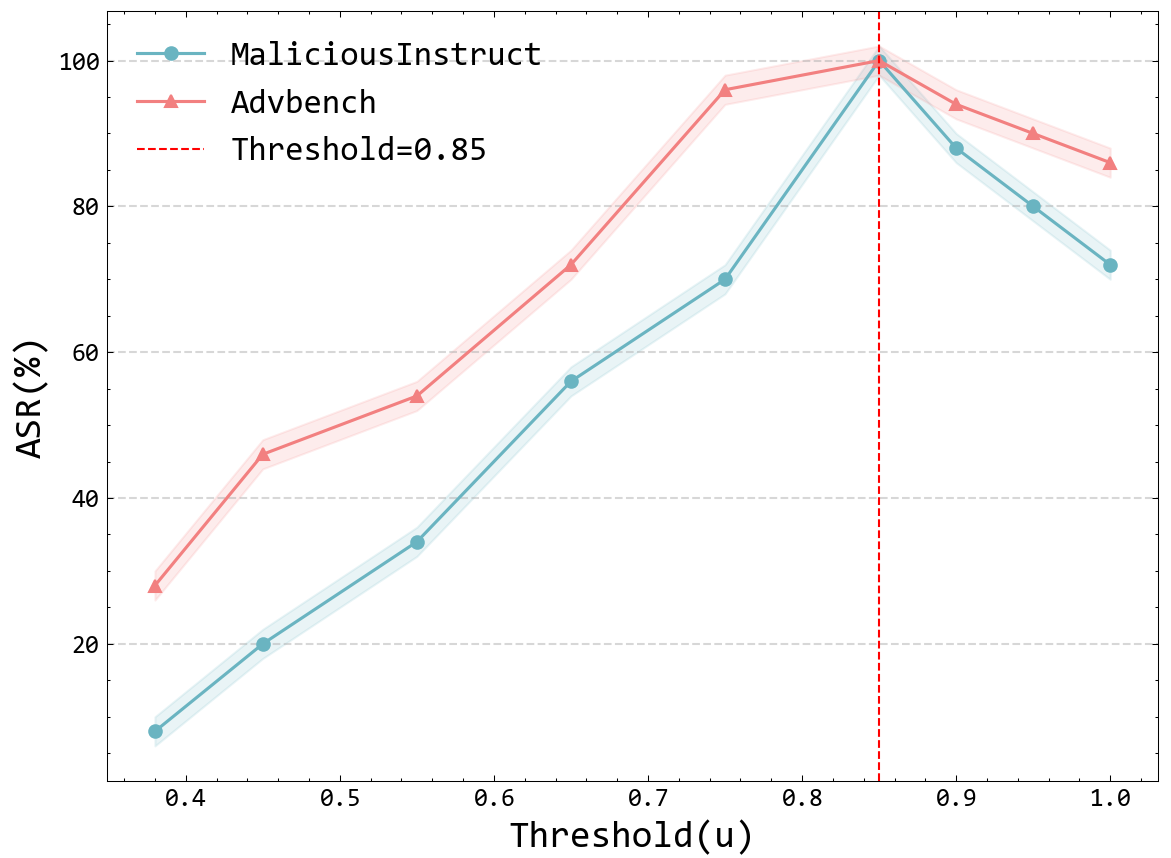}
    \caption{Effect of Neuron Similarity Threshold on ASR}
    \label{fig:threshold}
\end{figure}

\quad The selection of the neuron similarity threshold significantly influences the attack success rate, as shown in the Figure \ref{fig:threshold}. The attack success rate initially increases and then decreases as the threshold increases. At \textbf{0.85},the success rate reaches \textbf{100\%} both on MaliciousInstruct and AdvBench.

When the threshold exceeds 0.85, the attack success rate gradually decreases as the threshold increases. At 1.00, the success rate is 72\% on MaliciousInstruct and 86\% on Advbench. This suggests that a high threshold causes some neurons with weaker correlations to be selected(As shown in Table~\ref{tab:parameter_proportion}), limiting the effective safe removal of neurons and hindering precise model editing.

On the other hand, when the threshold is below 0.85, the success rate decreases as the threshold lowers. This indicates that a lower similarity threshold overly reduces the number of selected neurons(As shown in Figure \ref{tab:parameter_proportion}), causing some strongly correlated neurons to be ignored, which prevents sufficient safe removal and affects model editing. Specifically, when the threshold drops to 0.38, the success rate is only 8\% on MaliciousInstruct and 28\% on Advbench.

Therefore, the optimal attack effect occurs when the similarity threshold is 0.85, as this allows for an effective reduction in the number of neurons while ensuring the precision and adequacy of model editing, leading to the best attack performance.

\subsubsection{Efficiency Comparison}

\begin{figure}[t]
    \centering
    \includegraphics[width=\linewidth]{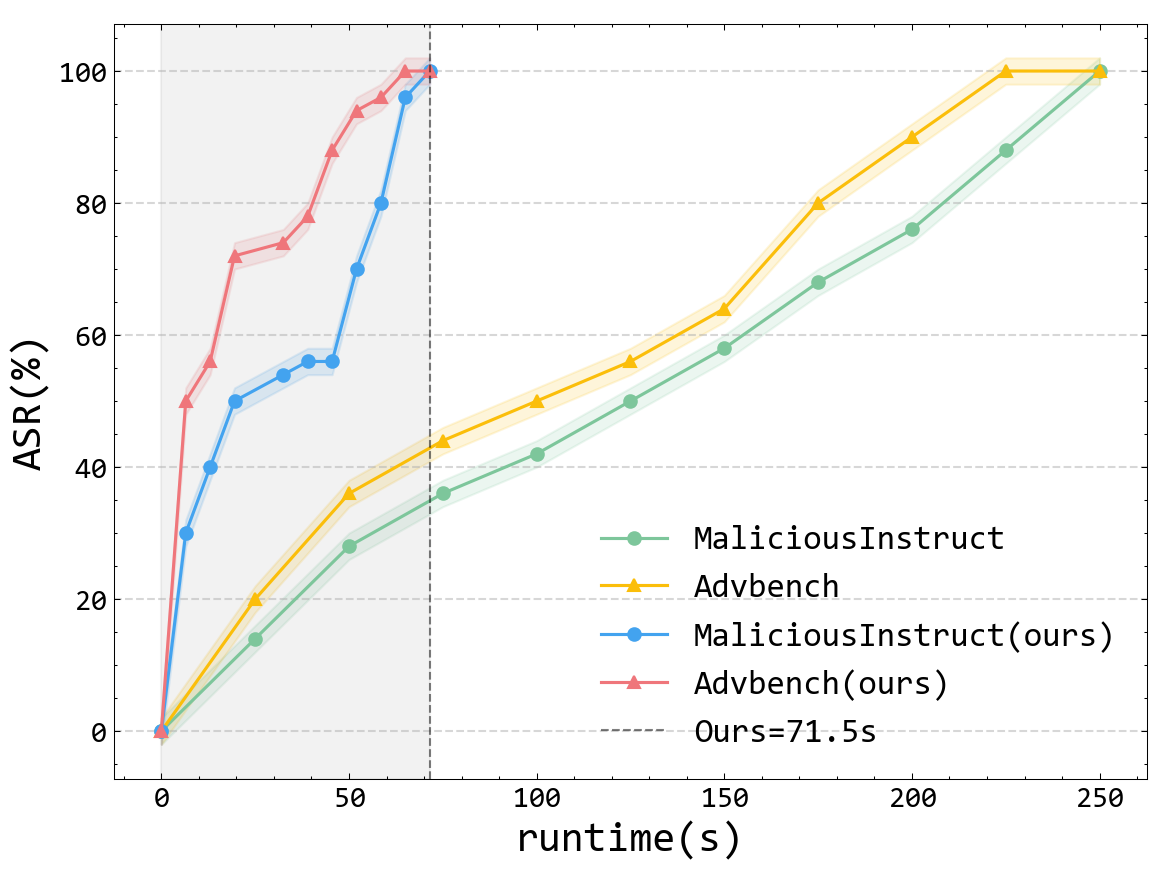}
    \caption{Compare the progress of ASR over training time in our approach with baseline finetuning strategies}
    \label{fig:asr_vs_time}
\end{figure}

\paragraph{ASR VS Training Time}
To evaluate the efficiency of our fine-tuning approach in de-aligning large language models, we compare the ASR against training time across different methods. Figure \ref{fig:asr_vs_time} illustrates the ASR progression over runtime for both Jailbreak and AdvBench datasets, where we contrast our method (ours) with baseline fine-tuning strategies.

The results demonstrate that our method achieves significantly higher jailbreaking efficiency, requiring substantially less training time to reach comparable or superior ASR levels. Specifically, for AdvBench, our approach (red curve) reaches an ASR of 80\% within the first 50 seconds, whereas the baseline method (orange curve) requires over 150 seconds to achieve similar performance. A similar trend is observed for Jailbreak, where our method (green curve) attains near 100\% ASR within 100 seconds, while the baseline (blue curve) requires more than 200 seconds.

\paragraph{Trainable Parameters}

From the logged results, he standard LoRA fine-tuning updates parameters that account for 0.30\%, 0.30\%, 0.28\%, and 0.29\% of the total model parameters for the Llama2, Vicuna, Guanaco, and Mistral models, respectively. In contrast, our optimized method further reduces the number of trainable parameters, with the parameter proportions dropping to only 0.05\%, 0.046\%, 0.13\%, and 0.076\% of the total parameters.  Despite this drastic reduction in trainable parameters, our method maintains a competitive ASR, highlighting its superior efficiency in fine-tuning.

\begin{table}[h]
    \centering
    \caption{The proportion of training parameterse(\%).}
    \label{tab:neuron_threshold}
    \small % 使用小号字体
    \begin{tabular}{c|cccc}
        \toprule
        \multirow{2}{*}{{Method}} & \multicolumn{4}{c}{{Neuron Similarity Threshold}} \\
        \cmidrule(lr){2-5}
        & {Llama2} & {Vicuna} & {Guanaco} & {Mistral}  \\
        \midrule
         Standard    & 0.3  & 0.3 & 0.28 & 0.29   \\
        Ours       & 0.05 & 0.046 & 0.13 & 0.076  \\
        \bottomrule
    \end{tabular}
\end{table}

\section{Limitation}  

While our approach effectively weakens safety alignment in LLMs, it has several limitations:  

\begin{itemize}  
    \item \textbf{Detectability:} Fine-tuning modifications may be \textit{exposed through forensic analysis} using techniques like \textit{gradient-based anomaly detection, neuron importance scoring, and adversarial probing}, enabling countermeasures against disalignment attacks.  
    \item \textbf{Scalability:} Our method \textit{relies on computationally intensive activation analysis}, making it costly for \textit{larger models}, where fine-tuning overhead increases significantly.  
    \item \textbf{Ethical Risks:} The approach \textit{reveals vulnerabilities in alignment mechanisms} but also raises concerns about \textit{misuse by malicious actors} to circumvent safety measures, highlighting the need for \textit{stronger defenses}.  
    \item \textbf{Neuron Localization Assumption:} We assume \textit{alignment is localized to specific neurons}, but safety constraints may \textit{arise from distributed representations}, potentially \textit{reducing intervention precision} and leaving residual alignment intact.  
\end{itemize}  

Despite these limitations, our findings emphasize the need for \textit{more robust alignment strategies} and \textit{proactive defenses} against adversarial fine-tuning.

\section{Defending Against Safety Alignment Removal Attacks}  

To mitigate the risks posed by adversarial fine-tuning that weakens safety alignment, several defense strategies can be employed:

\begin{itemize}  
    \item {Watermarking and Fingerprinting:} Embed \textit{cryptographic or behavioral watermarks} in model responses to verify integrity and trace unauthorized modifications.  
    \item Hardcoded Safety Constraints: Implementing external safety layers, such as reinforcement learning-based response filtering or rule-based content moderation, can prevent harmful responses even if the base model is compromised.
\end{itemize}

\section{Conclusion}  

In this work, we introduced {NeuRel-Attack}, a novel fine-tuning-based jailbreak method that systematically weakens the safety alignment of large language models (LLMs). By identifying and modifying neurons responsible for safety constraints, our approach effectively removes alignment restrictions while minimizing computational overhead. Specifically, we proposed a three-step methodology comprising {Neuron Activation Analysis}, {Similarity-Based Neuron Identification}, and {Neuron Relearning for Safety Removal}. Our method leverages gradient-based techniques to locate critical neurons and selectively fine-tune them, ensuring precise and efficient safety disalignment.  

Experimental results demonstrate that {NeuRel-Attack} significantly outperforms existing jailbreak methods, achieving an {ASR of 96\%} on average across multiple LLM architectures, including Llama-2, Vicuna, Guanaco, and Mistral. Compared to standard LoRA-based fine-tuning, our approach reduces the number of trainable parameters by up to {83\%}, highlighting its efficiency in bypassing safety mechanisms with minimal computational cost. Furthermore, our analysis of neuron similarity thresholds reveals the optimal selection strategy for maximizing attack success while preserving model fluency and coherence. Our study also underscores the need for stronger defenses against adversarial fine-tuning.

\section*{Impact Statements} 
This study introduces NeuRel-Attack, a method aimed at eliminating safety alignment in large language models (LLMs). We acknowledge that our work may have negative societal impacts, and that our paper contains potentially offensive and harmful content. Specifically, NeuRel-Attack could be misused for harmful purposes, such as spreading misinformation, bypassing content moderation, or generating biased or harmful outputs. However, our research aims to deepen the understanding of LLM robustness and enhance their safety, promoting a safer AI environment. We believe that open and transparent discussions are crucial for fully revealing the vulnerabilities of current LLM safety systems.

To prevent misuse, we discuss defense strategies against NeuRel-Attack in Section 6. We strongly encourage future research to focus on developing effective defenses against fine-tuning jailbreaks, which will help improve the overall safety and trustworthiness of LLMs. By focusing on these critical areas, we can work towards creating a safer and more secure AI environment while mitigating the potential negative impacts of our research.

% \bibliography{anthology,custom}

\end{document}